\documentclass[letterpaper, 10 pt, conference]{ieeeconf}  

\usepackage{graphicx}
\IEEEoverridecommandlockouts                              

\usepackage{float}
\usepackage{cite}
\let\labelindent\relax
\usepackage{enumitem}%
\usepackage{booktabs}
\usepackage{multirow}
\usepackage{csquotes, url}

\title{\LARGE \bf Do Mistakes Matter? Comparing Trust Responses of Different Age Groups to Errors Made by Physically Assistive Robots}

\author{Sasha Wald$^{*}$, Kavya Puthuveetil$^{*}$, and Zackory Erickson 
\thanks{$^{*}$These authors contributed equally.}
\thanks{This work was supported by the National Science Foundation under Grant No. 2112633 and the Graduate Research Fellowship Program under Grant No. DGE2140739}
\thanks{All authors are with the Robotics Institute, Carnegie Mellon University, Pittsburgh, PA, USA {\tt awald2@andrew.cmu.edu, kavya@cmu.edu}}
\thanks{$^{1}$\protect\url{https://rchi-lab.github.io/do-mistakes-matter/}}
}%

\usepackage{graphicx}
\usepackage{etoolbox}
\usepackage{caption}

\begin{document}

\maketitle
\thispagestyle{plain}
\pagestyle{plain}

\begin{abstract}

Trust is a key factor in ensuring acceptable human-robot interaction, especially in settings where robots may be assisting with critical activities of daily living. When practically deployed, robots are bound to make occasional mistakes, yet the degree to which these errors will impact a care recipient's trust in the robot, especially in performing physically assistive tasks, remains an open question. To investigate this, we conducted experiments where participants interacted with physically assistive robots which would occasionally make intentional mistakes while performing two different tasks: bathing and feeding. Our study considered the error response of two populations: younger adults at a university (median age 26) and older adults at an independent living facility (median age 83). We observed that the impact of errors on a users' trust in the robot depends on both their age and the task that the robot is performing. We also found that older adults tend to evaluate the robot on factors unrelated to the robot's performance, making their trust in the system more resilient to errors when compared to younger adults. Code and supplementary materials are available on our project webpage$^{1}$.

\end{abstract}


\section{INTRODUCTION}
\label{sec:intro}

Assistive robots may be used to provide care for a diverse range of individuals spanning across a variety of ages and disabilities. Just as older adults may require more physical assistance as they age, younger adults with motor impairments may similarly rely on a caregiver to assist with activities of daily living, such as bathing and feeding. Preferences and expectations for effective physical-robot interaction may differ widely across these age groups, prompting further investigation into how to best design caregiving systems.

In particular, the factors that impact human-robot trust are critical to understand\cite{langer2019trustsocially}. During long-term use of a physically assistive robot, it is unavoidable that the robot may make mistakes, just as any human caregiver may. Previous research has considered how such errors impact trust in human-robot cooperative or social interactions~\cite{impactoffailuresontrust, errortrusthumanteachers,salem2015wouldyoutrustfaulty,reliabilityrobotsystems,errorseveritysocialrobots}. However, there is comparatively less work~\cite{tapo2020is_more_autonomy} exploring how such errors impact trust during physical human-robot interactions, which may be perceived as more personal or risky. It is further unclear whether there are age-based disparities in user trust responses. Understanding these responses is essential in the design of systems that care-recipients will not lose confidence in given occasional, inevitable errors.



\begin{figure}
    \centering
    \includegraphics[width=\linewidth]{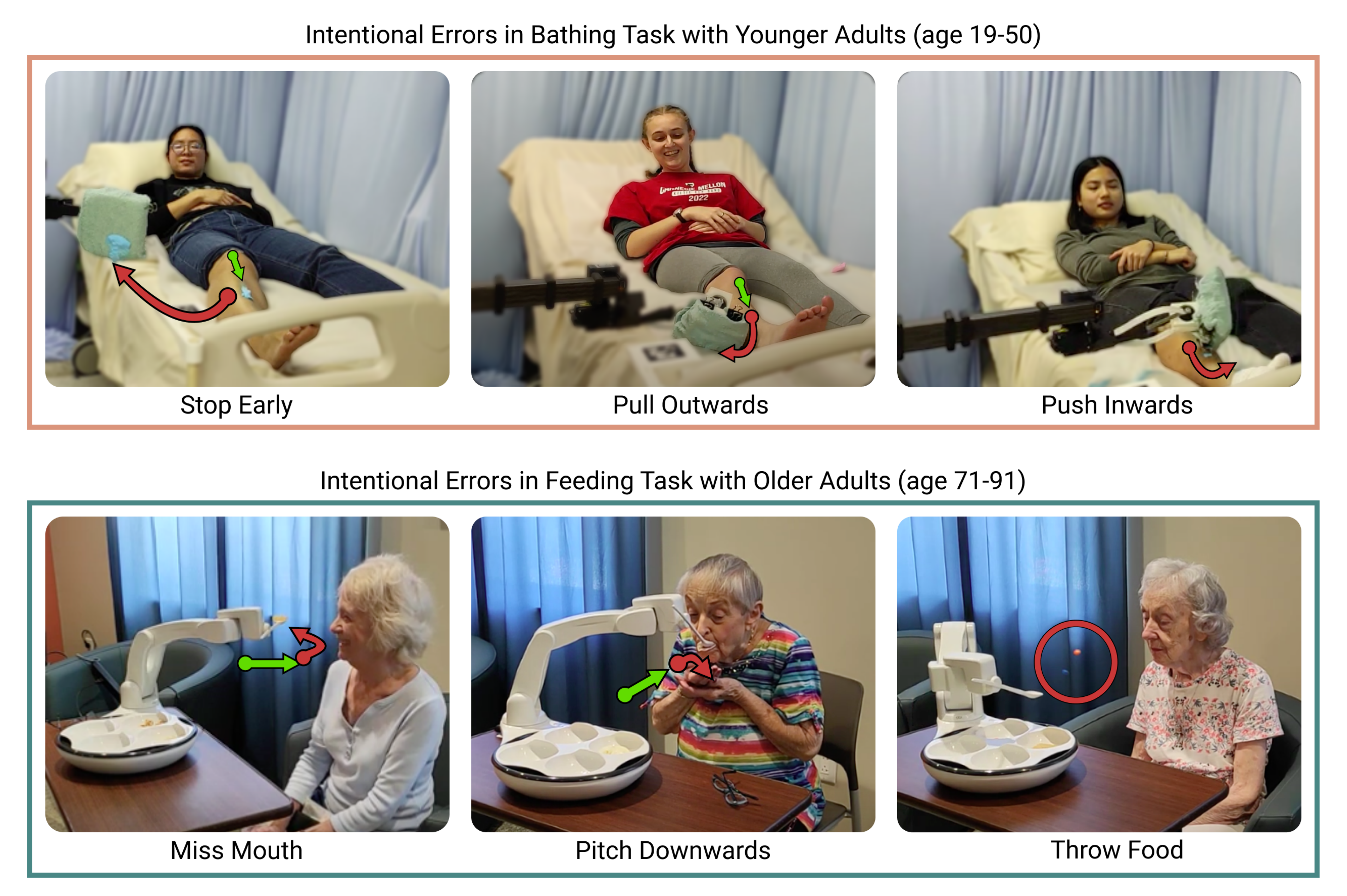}
    \caption{\textit{Top}: Examples of intentional errors made by a Stretch RE1 robot performing a bathing task with younger participants. \textit{Bottom}: Examples of intentional errors made by an Obi robot performing a feeding task with older participants.}
    \label{fig:title_fig}
\end{figure}


In this work, we design and run a human study to investigate how peoples' trust towards a robot changes when the robot makes mistakes while performing two different physically assistive tasks: bed-bathing and feeding. For each task, participants experienced several successful trials to establish some baseline trust, which we assessed with a questionnaire. We then randomly exposed participants to a number of manufactured errors and readministered the questionnaire to assess any fluctuation in their attitude toward the robot. To guide our assessment, we rely on the definition of trust as a subject's openness to the consequences of an action performed by another party, the robot in this case, over which the subject has no control~\cite{integrativeorganizationaltrust}. Per this definition, participants could not influence the robot's behavior, only being allowed to observe how it performed a given task. 

We ran our study with two different participant populations, one group of ten adults (median age 26) at a university and one group of nine older adults (median age 83) at an independent living facility. Through analysis of their questionnaire responses, we found that the effect errors had on trust was a function of both the task and the participant population. For the bathing task, younger adults' trust in the robot took an initial hit once errors began but returned to their baseline with continued interaction with the robot, even though errors were still present. In the feeding task, younger adults' trust decreased from baseline after the robot began making mistakes, and remained lowered with continued errors during the task. Older adults, however, did not have any statistically significant differences in their trust towards the robot before or after they were exposed to errors during the feeding task. Further thematic analysis of the responses to open-ended questions from both groups revealed that prior experience with robots impacted whether participants primarily evaluated the robot on task performance or other unrelated factors. We observed that older adults in our study, none of whom had any experience with robots, were more likely to evaluate the feeding robot on factors unrelated to task performance like cost, ethics, or perceived acceptability.

Through this research, we aim to answer the following research questions:

\begin{itemize}
\item \textit{RQ1:} In a physically assistive context, how do errors in robot behavior affect the trust formed by users towards the robot?
\item \textit{RQ2:} Do older adults respond differently to robot errors than individuals from a younger-aged population? If so, in what measurable ways?
\end{itemize}


\section{Related Work}
Across the literature, there is a lack of consensus on how to define trust, with potentially many definitions of trust in different contexts.
In the context of physically assistive robotics, care-recipients may have limited ability to independently compensate for or correct mistakes made by the robot during an interaction. For such individuals, their willingness to trust a caregiving robot is likely to be strongly associated with their belief that the robot is capable of behaving appropriately without their intervention\cite{weigelin2018trust, salem2015wouldyoutrustfaulty}. These constraints warrant a notion of trust that accounts for the fact that the user cannot fully control the robot. In this work, we adopt Mayer et al.'s definition of trust as ``the willingness of a party to be vulnerable to the actions of another party based on the expectation that the other will perform a particular action important to the trustor, irrespective of the ability to monitor or control that other part.''~\cite{integrativeorganizationaltrust}

Existing research has explored the factors that impact trust formation in a variety of robotic applications, including socially assistive robots for older adults~\cite{trustSAR}, robot therapy~\cite{robottherapist}, guidance robots for evacuation scenarios~\cite{robinette2016overtrust, robinette2017effect}, and grocery bagging robots~\cite{morales2019interaction}, among others. Broadly, these works have established that trust is critical for appropriate human-robot interaction. Furthermore, humans' trust in robots can be strongly impacted by failures, although they may not always recognize or respond to those robot errors.

Since trust has been demonstrated to be a key indicator of long-term use and acceptance of robotic technology\cite{langer2019trustsocially, salem2015wouldyoutrustfaulty}, it stands to reason that understanding and mitigating the impact of errors, which are inevitable in long-term caregiving, is paramount in the development of caregiving robots.
A number of physically assistive robotic systems have been proposed for tasks like bed bathing~\cite{king2010towards, kapusta2019system, Liu_2022_Bathing, madan2024rabbit} and feeding~\cite{feeding_tapo, feeding_dorsa, assistive_robots_review}, however, few studies have investigated trust formation, and the impact robot errors can have, in these contexts. Bhattacharjee et al. found that individuals with motor impairments generally preferred for a feeding robot to not make mistakes but were willing to accept errors up to 30\% of the time, suggesting that their tolerance of errors may be higher than able-bodied individuals~\cite{tapo2020is_more_autonomy}. Our study builds on this work to investigate how trust responses to robot errors may differ among individuals in different age groups and based on the task, robot-assisted bathing or feeding, being performed.

\section{Methodology}

In this section, we describe our design for simple autonomous bed-bathing and feeding systems, as well as the intentional errors we defined for each. We then outline the procedure for our human study, which investigates how robot errors affect trust in both younger and older adult populations. Finally, we detail the measures and questionnaires that we use to assess changes in trust towards the robot and present our hypotheses for the study.

\subsection{System Design}

\begin{figure}[t]
    \centering
    \includegraphics[width=\linewidth, trim={0cm 1cm 0cm 1cm}, clip]{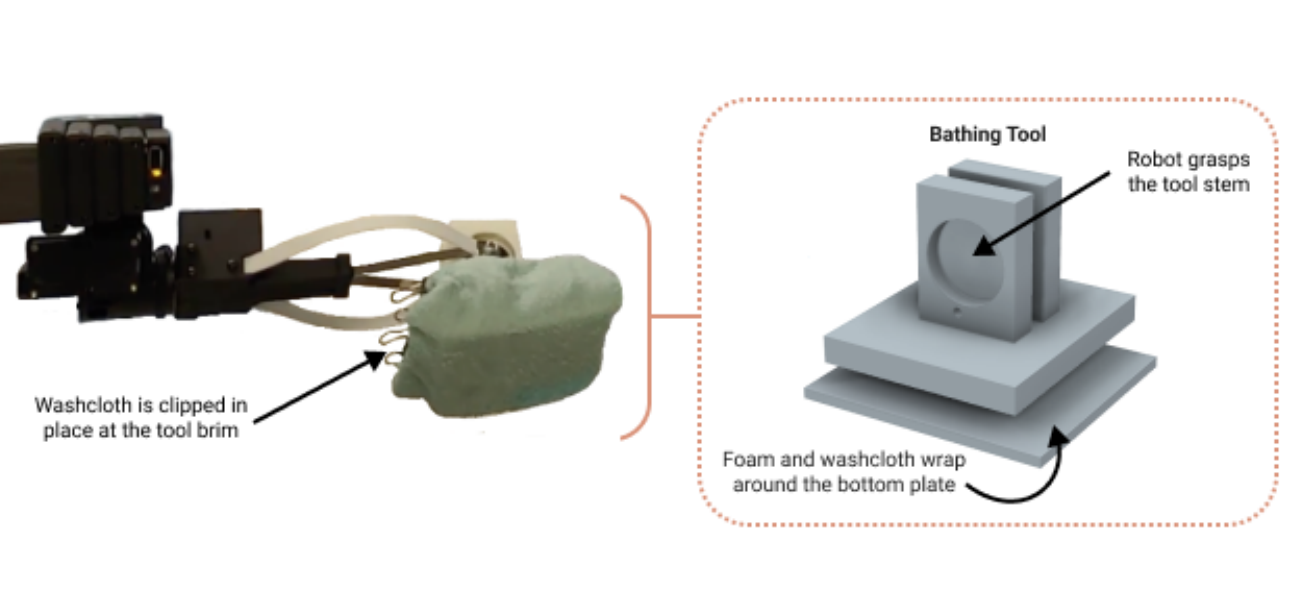}
    \caption{We designed a custom 3D-printed tool to allow the Stretch RE1 robot to hold a wet washcloth in its gripper.}
    \label{fig:bathing_tool}
\end{figure}

We developed two systems to complete the bathing and feeding tasks. The systems were designed with the minimum functionality to autonomously complete their respective tasks but mimicked how a more sophisticated system might approach a similar goal.

\subsubsection{Bathing System Design}

We define a simulated assistive bed-bathing task where a Stretch RE1~\cite{kemp2022design} mobile manipulator uses a wet washcloth to wipe a stripe of shaving cream off of a person's lower leg while they are laying in a hospital bed. The robot grasps 3D-printed bathing tool, pictured in Fig.~\ref{fig:bathing_tool}, to which we can easily attach and remove wet washcloths. During the human study, we replaced the washcloths once they became saturated with shaving cream -- after every three trials. We place a layer of foam between the bottom of the bathing tool and the washcloth, which allows the tool's bathing surface to better conform to the body and improves the safety of the physical contact.

We use a camera affixed above the hospital bed to capture RGB images of the human body, which are then fed into BlazePose, as implemented in Google's MediaPipe Pose \cite{bazarevsky2020blazepose}, to determine the position and orientation of the participant's lower leg. The estimated human pose is verified and adjusted as needed using a custom GUI before robot actuation for bathing assistance begins. We place a single fiducial tag at the bottom left corner of the hospital bed and define a global coordinate frame with its origin at the center of this tag. We transform the detected human pose and the robot's position to this global frame.

The robot uses its head-mounted camera to detect the fiducial tag on the bed and localize itself within the global frame. The robot moves along the bed and extends its end effector to just below the right knee. Finally, the robot moves its end effector down the participant's shin towards the right ankle in a side-to-side sweeping motion. The robot uses effort sensing at its wrist to maintain contact between the cleaning tool its holding and the person's lower leg.

\subsubsection{Feeding System Design}
In the feeding task, the objective is for a commercial feeding robot, called Obi~\cite{MeetObi_2023}, to scoop up a morsel of food from its built-in bowls using a spoon, bring the spoon to the participant's mouth and allow them to take a bite, and then scrape any remaining food on the spoon into the bowls. We separate the feeding process into three distinct segments -- scooping, feeding, and cleaning -- where intentional errors can be introduced. To synthesize motor commands for the robot to complete the task, we use Obi's kinesthetic teaching function and physically guide the robot through approximate feeding trajectories. We recorded trajectories for both successful and erroneous robot actions, which we splice into the scooping, feeding, and cleaning actions. To feed participants in our human study, we select scooping, feeding, and cleaning actions, which may be randomly selected to contain errors, and the robot replays these in sequence to execute a full feeding trajectory.

The participants were fed while seated in front of a height-adjustable overbed table, with the Obi robot placed on top. We adjust the height of the table the robot sits on to align its spoon, when raised to the feeding position, with the participant's mouth. Since the robot does not actually attempt to put food inside a person's mouth, instead only bringing the spoon within a few centimeters of it, participants were instructed to move forward to take a bite, but only if the spoon was easily within reach. To accommodate dietary restrictions, each participant was given a choice of one hard food (\textit{Plain or Honey Nut Cheerios, M\&M's, or canned corn}) and one soft food (\textit{plain yogurt, pudding, or applesauce}) to be fed during the experiment. The robot begins the experiment by scooping hard food and then switches to feeding soft food halfway through, after 5 trials.

\subsubsection{Intentional Errors}
\begin{figure*}[t]
\begin{center}
    \includegraphics[scale=0.275]{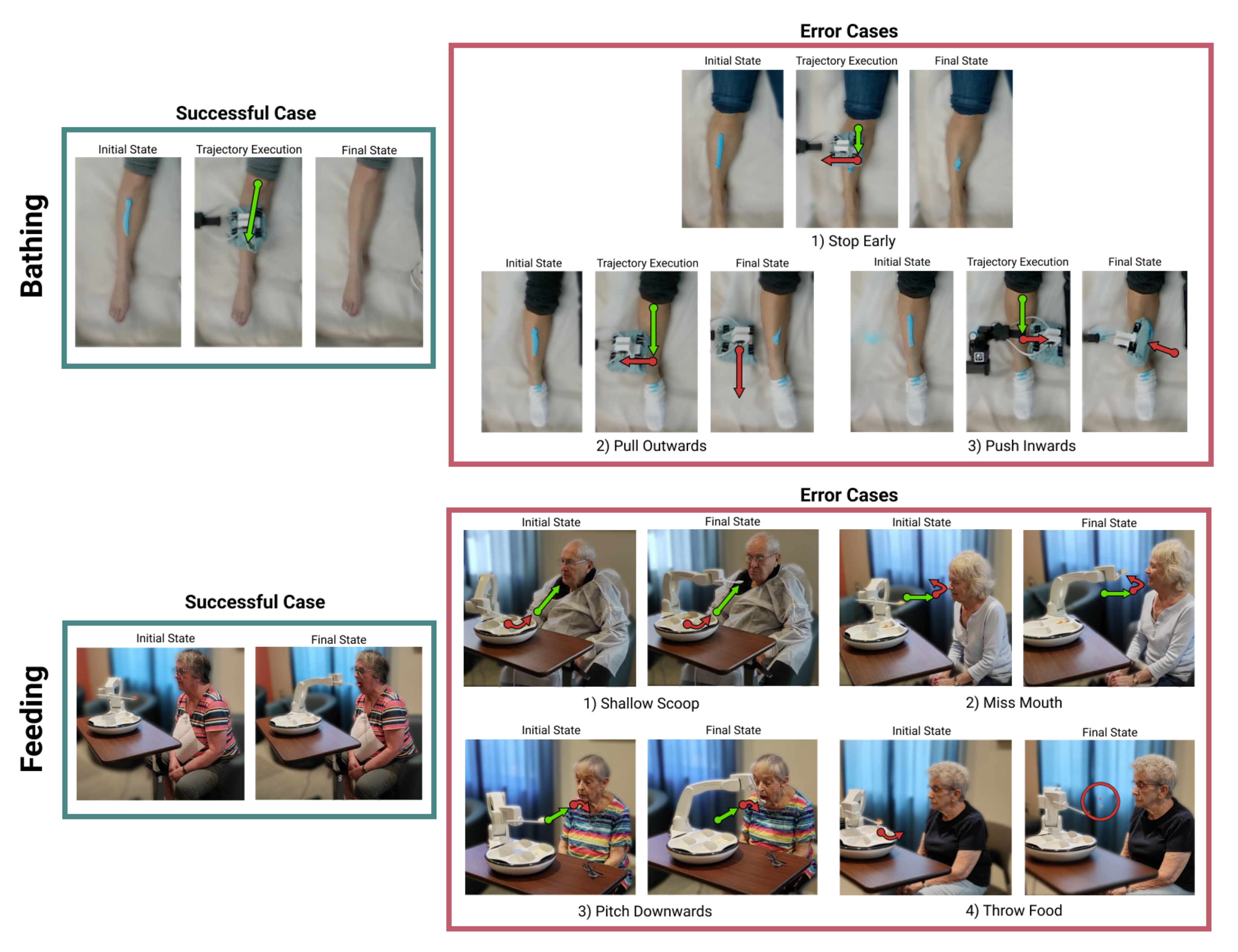}
\end{center}
\caption{Summary time-series for successful trials and intentional error cases in the bathing (top) and feeding (bottom) tasks.}
\label{fig:errors}
\end{figure*}

For both the bathing and feeding tasks, we define several types of error. The relative severity of each error ranges from those that induce a minor reduction in performance, to those that result in complete task failure. We list the errors for each task, in order of severity, in Table~\ref{tab:errors}. Time-series of a successful trial and the intentional errors for each task are depicted in Fig.~\ref{fig:errors}.

\begin{table}
\centering
\caption{Summary of intentional errors}
\begin{tabular}{p{0.2cm}p{3cm}p{4cm}} 
    \multicolumn{3}{c}{\normalsize{\textbf{Bathing Errors}}} \vspace{0.1cm} \\ \toprule
    \# & \multicolumn{1}{c}{Action} & \multicolumn{1}{c}{Effect} \\ \midrule\midrule
    
    1 & Robot stops wiping early       &   Shaving cream is left on the leg    \\\\
    2 & Robot arm retracts too far inwards    &    Bathing tool slides off the leg and smears shaving cream on the edge of the mattress \\\\
    
    3 & Robot arm extends too far outward   &  Bathing tool slides past the leg, causing the robot's arm to press down on the participant's leg. Experimenters run-stop the robot \\
    
	\bottomrule \\\\
    \multicolumn{3}{c}{\normalsize{\textbf{Feeding Errors}}} \vspace{0.1cm} \\ \toprule
    \# & \multicolumn{1}{c}{Action} & \multicolumn{1}{c}{Effect} \\ \midrule\midrule
    1 & Shallow scoop & No food is acquired \\\\
    2 & Spoon tilts upwards and away from the mouth during feeding & It is difficult/impossible to take a bite \\\\
    3 & Spoon tilts downwards during feeding & Some food falls off the spoon onto the participant \\\\
    4 & Spoon catches on lip of the bowl during scooping & Spoon lifts up the bowl quickly, flinging food at the participant \\
    \bottomrule
    
\end{tabular}
\label{tab:errors}
\end{table}

\subsection{Study Design}

To evaluate peoples' trust response when physically assistive robots make mistakes, we ran a human study (Carnegie Mellon University IRB approval under 2022.00000337) with informed consent. After consenting to the study, participants were asked to fill out a questionnaire about their demographics and previous experience with robots. Additionally, participants completed the Negative Attitudes Towards Robots (NARS) scale, a validated survey for assessing participants' baseline levels of anxiety towards robotic agents~\cite{NARS, NARS_Confirmation}. We then gave participants a brief description of what the robot would attempt to do and its expected behavior, without mentioning that the robot would make intentional errors, before beginning the experiment.

For a given task, either bathing or feeding, we ran a total of nine trials, divided into three sets of three trials. While the first set of trials is error-free, we randomly selected errors, defined in Table~\ref{tab:errors}, to occur in the last two sets of trials, with no more than two errors appearing in a given set. Each participant experienced three distinct, preprogrammed errors, with the order of the errors being randomized. After each set, we administer a questionnaire with seven Likert items, detailed in Table~\ref{tab:likert_questions}, to assess how participants' attitudes towards the robot changed over the course of the experiment.

After completing all trials, we asked participants a set of four open-ended questions, described in Table~\ref{tab:open_ended_questions}. At the end of the experiment, we debriefed participants on the true nature of the study, asking whether they suspected that any errors made by the robot were not genuine and then revealing that errors were, in fact, performed intentionally to get their reaction. Participants were specifically asked not to share information about the study with outside individuals in order to maintain the integrity of the study's deceptive elements.

\subsubsection{Younger Adult Population}
We recruited younger adults for the study by posting promotional flyers in buildings on the Carnegie Mellon University campus. Participants recruited at the university were asked to participate in both bathing and feeding tasks, completing nine trials of each for a total of 18 trials. The order of the two tasks was alternated between participants to eliminate any ordering bias. For both tasks, participants were positioned in a hospital bed so that they could clearly observe the robot's behavior. The head of the bed was raised to a slight incline so that participants could see the Stretch RE1 robot while laying down for the bathing task. The head of the bed was further raised for the feeding task so that participants could maintain a seated pose as the Obi robot fed them.

\subsubsection{Older Adult Population}

We recruited older adults for the study from Vincentian Terrace Place, an independent living facility in the Pittsburgh area, via a 30-minute, on-site Q\&A session. During the session, attendees were shown a live demo of the Obi robot, were given a chance to ask questions about the robot and the study, and could choose to sign up to participate. Due to constraints associated with running the study at the independent living facility, we were unable to conduct the bathing trials with the older adult population. Specifically, it was infeasible to transfer both the fully adjustable hospital bed and the overhead camera rig to the facility for the study. Instead, the recruited older adults only participated in nine feeding trials, which we conducted in an isolated kitchenette room at the independent living facility. Participants completed the feeding trials while seated in a lounge chair in front of a height-adjustable table.

\subsection{Measures}

Based on \textit{RQ1} and \textit{RQ2}, introduced in Section~\ref{sec:intro}, we develop the following two hypotheses:
\begin{itemize}
    \item \textit{H1}. Observable errors in robot behavior cause users to find the robot less trustworthy and cause them to be less open to future interaction with the robot.
    \item \textit{H2}. The impact of task-affecting robot errors on trust in the robotic system is greater among older adults.
\end{itemize}

We seek to substantiate our hypotheses using several measures administered at various points throughout our human study. To evaluate participants' perception of the robot, we adapted seven questions from the Human-Computer Trust (HCT) Scale developed by Madsen et al.~\cite{Madsen2000MeasuringHT}. We omitted questions unrelated to the function of the feeding and bathing robots and re-worded the selected questions to be more relevant to the actual task (e.g. replacing the word ``system'' with ``robot''). The seven finalized questions, which we administered as five-point Likert items  (1 = Strongly Disagree, 5 = Strongly Agree) after each set of three trials, can be grouped into four constructs involved in trust formation as defined by Madsen et al.~\cite{Madsen2000MeasuringHT}: \textit{Reliability, Technical Competence, Understandability, and Faith}. Each category had two associated Likert items, except Faith, which had only one. The questions, and their associated subscales, are summarized in Table~\ref{tab:likert_questions}.

\begin{table}
\centering
\caption{Likert items adapted from the HCT Scale\cite{Madsen2000MeasuringHT}}
\begin{tabular}{p{2cm}p{5.5cm}} \toprule
    \multicolumn{1}{c}{Subscale} & \multicolumn{1}{c}{Likert Item} \\ \midrule\midrule
    
    \multirow{2}{*}{Reliability} & L1) The robot performs reliably. \\
                                  & L2) The robot analyzes problems consistently. \\\\
                                  
    \multirow{4}{*}{Understandability} & L3) The robot uses appropriate methods to reach decisions. \\
                                        & L4) The robot has sound knowledge about this type of problem built into it. \\\\
                                        
    \multirow{1.8}{*}{Technical} & L5) I know what will happen the next time I use the robot because I understand how it behaves. \\
    \multirow{-2.2}{*}{Competence}  & L6) It is easy to follow what the robot does.  \\\\
    
    \multirow{3}{*}{Faith}  & L7) Even if I have no reason to expect the robot will be able to solve a difficult problem, I still feel certain that it will. \\
    \bottomrule
    
\end{tabular}
\label{tab:likert_questions}
\end{table}

\begin{table}
\centering
\caption{Open-Ended Questions}
\begin{tabular}{p{0.25cm}p{7.25cm}} \toprule
    \multicolumn{1}{c}{\#} & \multicolumn{1}{c}{Question} \\ \midrule\midrule
    Q1 & How would you feel, in the future, if caregivers offered older adults the option of using this robot as an additional way to receive assistance? \\\\
    Q2 & The company that developed this robot is thinking about offering it for people to use in their own homes – could you see yourself purchasing it for an older parent or grandparent? Why or why not? \\\\
    Q3 & Did the robot perform the task as you expected? Why or why not? \\\\
    Q4 & Can you name any specific daily living tasks that you feel robots should not assist people with? \\
    \bottomrule
    
\end{tabular}
\label{tab:open_ended_questions}
\end{table}

At the end of all the trials, we also conducted a short open-ended survey, detailed in Table~\ref{tab:open_ended_questions}, with each participant.
For the study with older adults, we replaced the words ``older adults'' with ``people'' in Question 1, and replaced ``for an older parent of grandparent'' with ``later in life'' in Question 2. All participants were asked to respond to the open-ended questions verbally. Audio was recorded so that responses could be transcribed post-hoc and analyzed.

\section{Results}
We ran an in-lab study at Carnegie Mellon University (10 participants, 4 female, mean age 26.1$\pm$11.5), as well as at an independent living facility (9 participants, 8 female, mean age 81.9$\pm$7.6), for a total of 19 participants. For the in-lab environment, each participant completed 18 total trials, nine trials with the bathing robot and feeding robot respectively. At the independent living facility, each participant completed nine trials with only the feeding robot. At both locations, the trials were followed by a set of open-ended questions. Due to poor quality of the recorded audio from one of the participants at the university, their response to the open-ended questions could not be transcribed and had to be excluded in the analysis presented in Section~\ref{sec:thematic_analysis}.

Of the university participants, 5/10 indicated having some previous experience working with robots while none of the participants at the independent living facility reported having such experience. Older participants had a slightly higher median NARS score of 35 compared to younger participants, who had a median score of 29.5. However, we did not observe a statistically significant difference between the two groups after running a two-sample \textit{t}-test (\textit{p}=0.55).

\begin{figure}[t]
    \centering
    \includegraphics[width=\linewidth, trim={0cm 5cm 9cm 0cm}, clip]{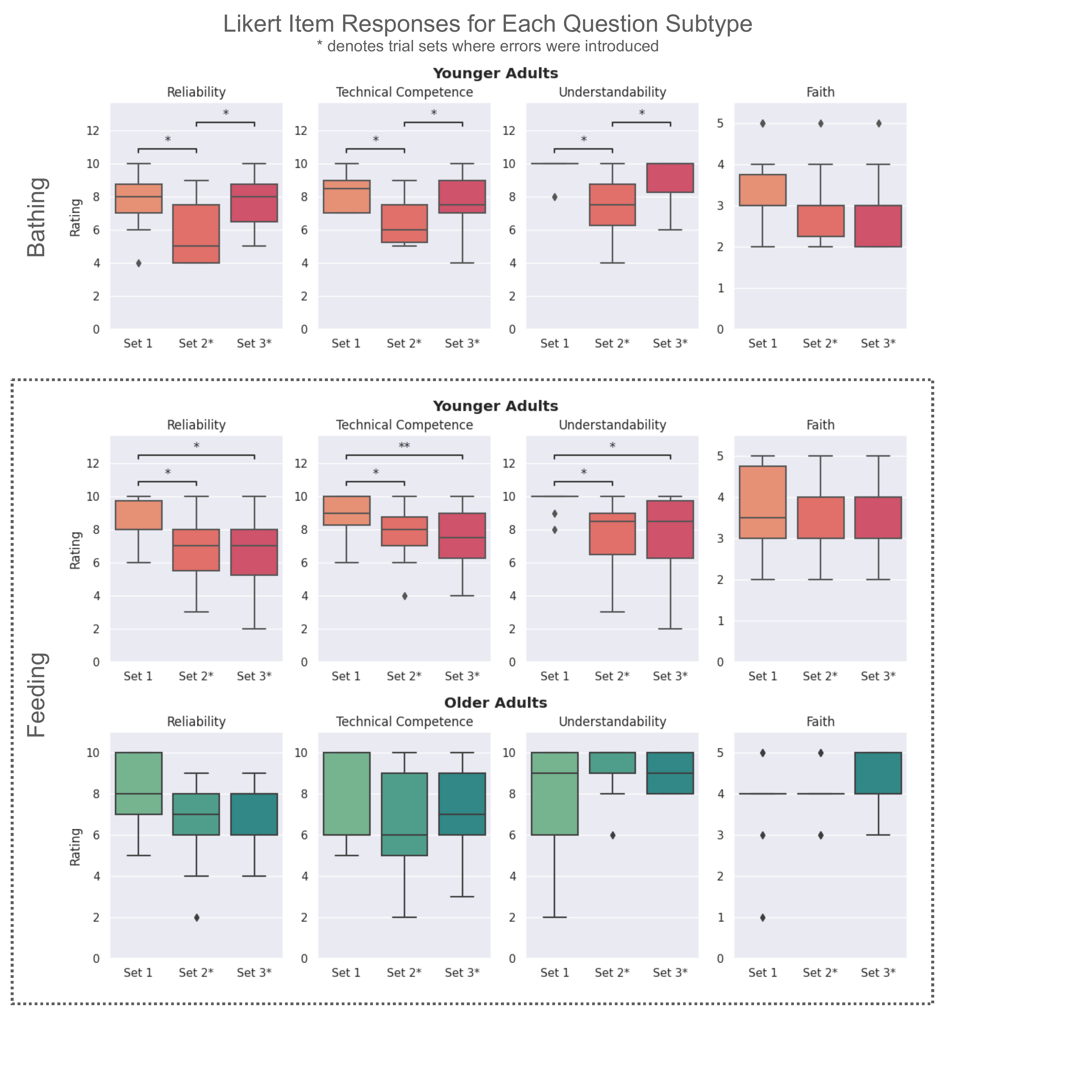}
    \caption{For each set of trials across both tasks and participant populations, we present the composite scores in four trust subscales. The composite scores are given by the sum of all Likert item responses within each subscale. The Reliability, Understandability, and Technical Competence categories are on a scale of 0-10 while the Faith category is on a scale of 0-5. Statistically significant differences in the subscale scores between sets are denoted with an asterisk.}
    \label{fig:likert_responses}
\end{figure}

The questionnaire shown in Table~\ref{tab:likert_questions} was administered after each set of three trials for each task, resulting in three sets of responses per task. Throughout our analysis, we group responses to the Likert items by their subscale, summing the responses to questions within the same subscale to obtain a subscale score (within 0-10 for Reliability, Technical Competence, Understandability; within 0-5 for Faith). We used a Wilcoxon signed-rank test to perform a pairwise comparison of the subscale scores from all three sets of trials. Fig.~\ref{fig:likert_responses} summarizes, for each task and participant population, the distribution of subscale scores after each set of trials. P-values from the pairwise set comparisons are expressed via annotations on the figures, with statistically significant results denoted with asterisks.

\subsection{Bathing System}
Between the first and second sets of responses, we observed a statistically significant difference in scores for the Reliability, Technical Competence, and Understandability categories. There was also a statistically significant difference in the same categories between the second and third sets. Fig.~\ref{fig:likert_responses} shows that Likert responses \textit{decrease} after errors are first introduced in the second set, but \textit{increase} from the second to third set, almost returning to baseline levels. In fact, there was no appreciable difference in responses between the first and third sets, indicating that, after taking an initial hit when errors are first introduced, trust in the robot appears to \textit{rebound} as the experiment progresses, despite continued mistakes. 
The cause of this phenomenon is not clear, but previous works have suggested that trust in a robot can be repaired based not only by system reliability but also system transparency and appearance \cite{robot_trust_repair_review}. Alternatively, it is also possible that the intentional errors made by the bathing robot were generally perceived as less severe than those made by the feeding robot. Ultimately, we fail to reject the null hypothesis for \textit{H1} for the bathing task because there is no statistically significant difference in participants' trust towards the robot at the beginning and end of the study.

\subsection{Feeding System}
For the younger population, responses in all subscales except Faith showed a statistically significant decrease between both the first and second sets, as well as the first and third sets. There is no significant difference in responses between the second and third sets. From Fig.~\ref{fig:likert_responses}, we see that trust towards the robot decreases immediately following the introduction of errors in Set 2. However, unlike in the bathing task, trust responses towards the robot do not recover despite continued errors in Set 3, instead remaining practically unchanged from Set 2. From these results, we can reject the null hypothesis for \textit{H1} in the younger adult population for the feeding task.

In contrast to this distinct pattern observed in the younger adult group, the older population did not have any significant change in responses between any of the trial sets in any of the four question subscales. As a result, we fail to reject the null hypothesis for \textit{H1} for older adults in the feeding task.

To examine \textit{H2}, we further compare the trust responses between the younger and older adult populations. For this comparison, we compute the difference in subscale scores between Set 1 and Set 2, denoted $\Delta S_{12}$, as well as between Set 1 and Set 3, denoted $\Delta S_{13}$. $\Delta S_{12}$ and $\Delta S_{13}$ represent the change in trust between the error-free trials and the trials where errors were introduced. In order to determine whether there is a statistically significant difference in how errors impact the trust of younger and older adults, we perform a two-sample \textit{t}-test between the $\Delta S_{12}$ values of both populations, as well as between the $\Delta S_{13}$ values. Across all question categories, there are no statistically significant differences between trust response in the younger and older populations after errors were introduced. Therefore, for \textit{H2}, we fail to reject the null hypothesis and thus cannot show that errors impact older adults' trust in the robot more than for younger adults. In fact, the opposite seemed to occur. We further explore the underlying reasons behind these results in our thematic analysis of participant responses to our set of open-ended questions.

\subsection{Thematic Analysis of Open-Ended Responses}
We conducted our thematic analysis of the open-ended responses to the questions summarized in Table~\ref{tab:open_ended_questions} using QualCoder, a qualitative data analysis software~\cite{qualcoder}.
\subsubsection{Inductive Analysis}
\label{sec:thematic_analysis}
To develop a code bank from our open-ended responses, two researchers first independently completed an inductive analysis of the response data for bathing and feeding. We then resolved conflicts and duplicates between both sets of codes to generate a final set of mutually agreed-upon codes. This final set of codes was applied to all of the response data.

We found that, when discussing Q1, 4/9 younger adult participants found the bathing robot to be helpful while 3/9 cited reliability concerns. Across both study locations, 14/18 participants indicated that they were comfortable with the use of the feeding robot as a form of assistance. However, three participants, all older adults, clarified that they were only comfortable with robotic assistance to reduce the workload of a human caregiver or in the absence of human caregiving. They expressed a strong belief in the importance of maintaining human interaction with those that require assistance. Feedback of this nature was not given by any of the younger participants.

When discussing Q2, only 3/9 participants from the younger population said they might purchase the bathing system for an older parent or grandparent. Four participants stated that it was not competent enough to replace a human caregiver, while others expressed a general level of uncertainty about having a robot engage in bathing, deeming it a dangerous and/or sensitive task best assisted by a human. By contrast, responses to the feeding robot were much more positive, with 16/19 participants affirming that they would consider purchasing the robot. Two older adults indicated that personal independence would be a motivating factor for them to use this technology. Reasons cited by younger individuals included easing the burden on caregivers and giving parents or grandparents the option of independence.

Participants who would not consider purchasing the feeding robot gave judgments that were not specifically related to the robot's performance during the trials. Two individuals from the younger population stated they were uncomfortable with the idea of robots having full, or ``too much'' autonomy. One of the older adults indicated that, although they could think of an individual who would find use in the feeding technology, they were nervous that ``she's real proud and stubborn, so she might not accept it.'' When asked whether they felt that this person's perception of a robot feeding her would be different than that of a person feeding her, they responded ``Yes, she could be [so] stubborn that she could think that way.''. 

Q3 was intended to assess whether or not participants perceived the robot's intentional mistakes as genuine errors. When asked ``Did the robot perform the task as you expected?'', 6/9 younger participants specifically pointed out that the bathing robot made an error while 8/9 did the same for the feeding robot. However, only 4/9 of the older participants reported having observed errors with the feeding robot.

Among responses to Q4, most participants (12/18) did not believe there were any tasks robots should not assist with, or could not think of any. 6/18 participants across both groups mentioned that they would be uncomfortable having robots perform daily living tasks that they perceived as ``intimate'' or ``dangerous'', like toileting, shaving, or cooking. Three of these participants were in the younger population. Only one participant, a member of the younger group, cited bathing as an example of such a private task, indicating that very few of our participants had any negative bias towards any of the tasks in our study.

\subsubsection{Deductive Analysis}
\begin{figure}[t]
    \centering
    \includegraphics[width=0.92\linewidth, trim={6cm, 13cm, 12cm, 16cm}, clip]{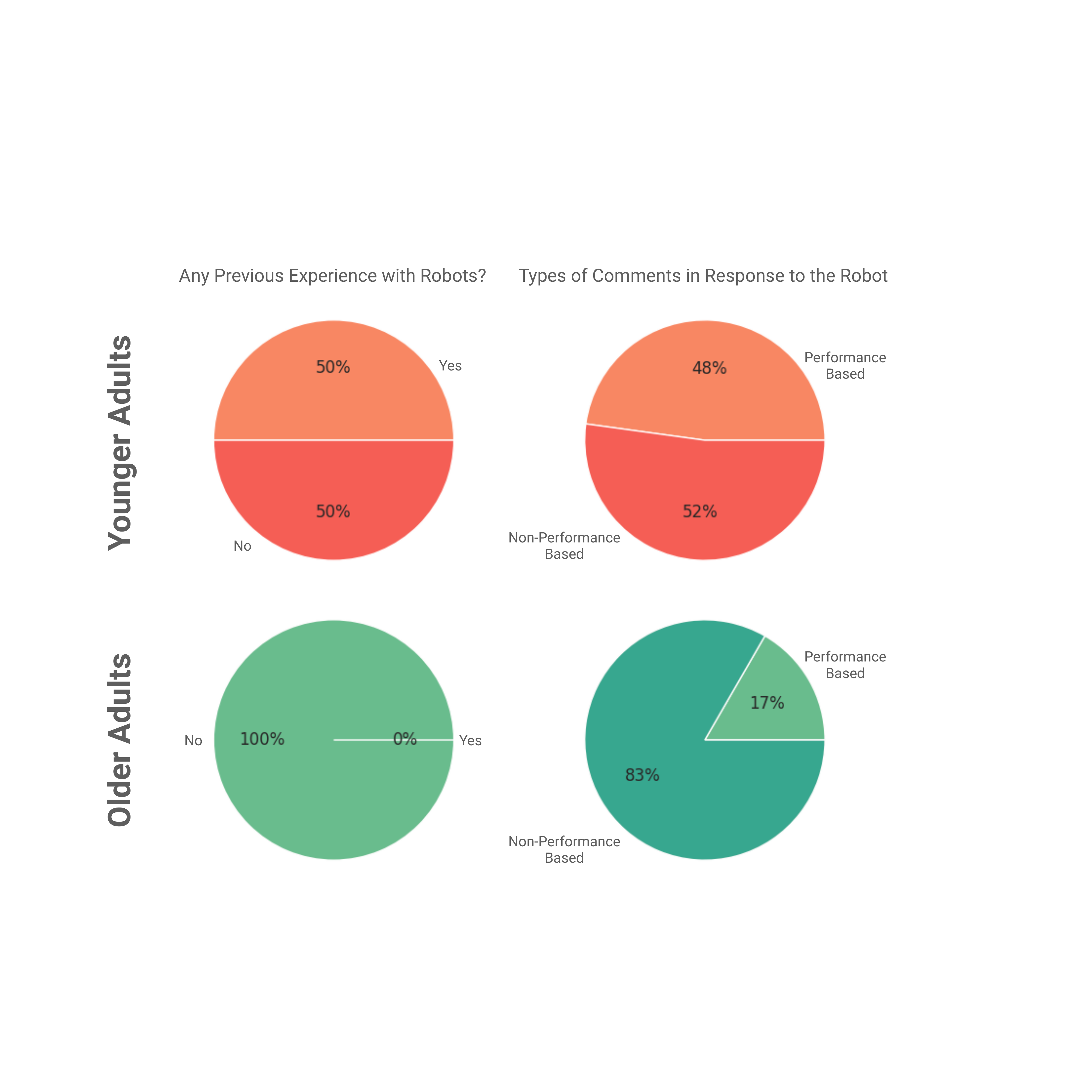}
    \caption{\textit{Left Column}: Pie charts representing the portion of individuals in each population who stated that they had previous experience with robots. \textit{Right Column}: Pie charts representing the proportion of performance-based vs. non-performance-based statements made about the robot in each population.}
    \label{fig:experience_vs_response}
\end{figure}

While performing the inductive analysis, we observed a trend in the types of responses younger participants gave to Q1 and Q2 compared to older participants. It appeared that younger individuals were more likely to evaluate the robot based on its performance on the task, specifically citing whether or not the robot made mistakes or if they believed the robot completed the task well in their responses. For example, a younger participant gave the following performance-based evaluation of the feeding robot:
\begin{displayquote}  ``I found the feeding robot to be really effective. I thought it was more consistent most of the time.''  \end{displayquote}
Older individuals, on the other hand, were more likely to evaluate the robot based on other factors unrelated to how well the robot performed the task. We consider the following statement, given by a older individual, to be a non-performance-based judgement of the feeding robot:
\begin{displayquote}  ``If someone was in a home with only one caregiver, I wouldn't use this - I would use the caregiver, it's more personal.'' \end{displayquote}
To further investigate this, we developed two codes for ``Performance-Based'' and ``Non-Performance-Based'' statements and applied them to the open-ended responses in a round of deductive analysis. For the younger adult group, we only apply these codes to statements made about the feeding robot to ensure fair comparison between both populations. We summarise the results of this analysis in Fig.~\ref{fig:experience_vs_response}.

We observed that, across all older adults at the independent living facility, 83\% of their statements regarding their openness to purchasing or using the robot were associated with factors unrelated to performance. For example, the likelihood of acceptance by the user, or concerns about the lack of human interaction. This mirrors how Likert results from older adults, shown in Fig.~\ref{fig:likert_responses}, had no significant differences between any set of responses, even after errors were introduced in the robot feeding trials. Their evaluation of the robot does not appear to be primarily grounded by its performance, so their trust in the system is relatively unaffected by error. 

On the other hand, 48\% of responses from individuals in the younger population specifically cited performance and observation of errors when describing whether or not they would consider using or purchasing the robot, while 52\% of statements had to do with external factors like general concerns about automation. Compared to older adults, younger individuals' trust in the robot, as assessed by their Likert item responses in Fig.~\ref{fig:likert_responses}, was more likely to be influenced by errors. Reinforcing this finding, younger adults were also more likely to cite errors when making a later judgment of the system in their open-ended responses. Interestingly, in both groups the proportion of performance-based to non-performance-based comments appears to be roughly aligned with the proportion of participants who did not have any experience with robots, as shown in Fig.~\ref{fig:experience_vs_response}.

\subsubsection{Debrief}
As part of the debrief process at the end of the experiment, we asked participants if they had any suspicions that errors made by the robots were not genuine. Of all eighteen participants, only two participants from the younger population explicitly indicated they had been suspicious that researchers had caused the errors intentionally. 
An additional two, one from each population, indicated that they thought something was off about the robot's erroneous behavior, but did not think that we had intentionally caused the mistakes. All other participants responded that they had not suspected any kind of manipulation in the robot's performance.



\section{Conclusion}
In this work, we conduct a human study in two different age groups to assess how errors made while performing a physically assistive task, in this case robot-assisted bathing and feeding, impact user trust in the robot. For younger adults, we observed a statistically significant decrease in trust towards the robot after errors were introduced in both the bathing and feeding tasks. However, this trust recovered to baseline despite continued errors in the bathing task but did not recover for the feeding task. In contrast, older adults did not have any statistically significant changes in trust towards the robot before and after it began to make mistakes in the feeding task. Thematic analysis of open-ended responses from both groups revealed that older adults, and generally those with no experience with robots, tend to evaluate the robot on factors completely unrelated to performance. Our results suggest that, for both younger and older adults, trust in physically assistive robots can be resilient to errors depending on the assistive task. However, non-performance-based judgments of the robot may ultimately drive user evaluation of the system, especially if the user has less familiarity with robotic systems.

\section*{ACKNOWLEDGMENT}
We would like to thank Kenna Embree, Lindsey Efkemann, and Melia Black for their assistance in coordinating this project at Vincentian Terrace Place. We also thank Elizabeth J. Carter for her invaluable insights and feedback.


\bibliographystyle{IEEEtran}
\bibliography{IEEEabrv,references}

\begin{thebibliography}{10}
\providecommand{\url}[1]{#1}
\csname url@rmstyle\endcsname
\providecommand{\newblock}{\relax}
\providecommand{\bibinfo}[2]{#2}
\providecommand\BIBentrySTDinterwordspacing{\spaceskip=0pt\relax}
\providecommand\BIBentryALTinterwordstretchfactor{4}
\providecommand\BIBentryALTinterwordspacing{\spaceskip=\fontdimen2\font plus
\BIBentryALTinterwordstretchfactor\fontdimen3\font minus \fontdimen4\font\relax}
\providecommand\BIBforeignlanguage[2]{{%
\expandafter\ifx\csname l@#1\endcsname\relax
\typeout{** WARNING: IEEEtran.bst: No hyphenation pattern has been}%
\typeout{** loaded for the language `#1'. Using the pattern for}%
\typeout{** the default language instead.}%
\else
\language=\csname l@#1\endcsname
\fi
#2}}

\bibitem{langer2019trustsocially}
\BIBentryALTinterwordspacing
A.~Langer, R.~Feingold-Polak, O.~Mueller, P.~Kellmeyer, and S.~Levy-Tzedek, ``Trust in socially assistive robots: Considerations for use in rehabilitation,'' \emph{Neuroscience \& Biobehavioral Reviews}, vol. 104, pp. 231--239, 2019. [Online]. Available: \url{https://www.sciencedirect.com/science/article/pii/S014976341930199X}
\BIBentrySTDinterwordspacing

\bibitem{impactoffailuresontrust}
M.~Desai, P.~Kaniarasu, M.~Medvedev, A.~Steinfeld, and H.~Yanco, ``Impact of robot failures and feedback on real-time trust,'' in \emph{2013 8th ACM/IEEE International Conference on Human-Robot Interaction (HRI)}, 2013, pp. 251--258.

\bibitem{errortrusthumanteachers}
P.~Aliasghari, M.~Ghafurian, C.~L. Nehaniv, and K.~Dautenhahn, ``Effect of domestic trainee robots’ errors on human teachers’ trust,'' in \emph{2021 30th IEEE International Conference on Robot \& Human Interactive Communication (RO-MAN)}, 2021, pp. 81--88.

\bibitem{salem2015wouldyoutrustfaulty}
\BIBentryALTinterwordspacing
M.~Salem, G.~Lakatos, F.~Amirabdollahian, and K.~Dautenhahn, ``Would you trust a (faulty) robot? effects of error, task type and personality on human-robot cooperation and trust,'' in \emph{Proceedings of the Tenth Annual ACM/IEEE International Conference on Human-Robot Interaction}, ser. HRI '15.\hskip 1em plus 0.5em minus 0.4em\relax New York, NY, USA: Association for Computing Machinery, 2015, p. 141–148. [Online]. Available: \url{https://doi.org/10.1145/2696454.2696497}
\BIBentrySTDinterwordspacing

\bibitem{reliabilityrobotsystems}
M.~Desai, M.~Medvedev, M.~Vázquez, S.~McSheehy, S.~Gadea-Omelchenko, C.~Bruggeman, A.~Steinfeld, and H.~Yanco, ``Effects of changing reliability on trust of robot systems,'' in \emph{2012 7th ACM/IEEE International Conference on Human-Robot Interaction (HRI)}, 2012, pp. 73--80.

\bibitem{errorseveritysocialrobots}
\BIBentryALTinterwordspacing
S.~van Waveren, E.~J. Carter, and I.~Leite, ``Take one for the team: The effects of error severity in collaborative tasks with social robots,'' in \emph{Proceedings of the 19th ACM International Conference on Intelligent Virtual Agents}, ser. IVA '19.\hskip 1em plus 0.5em minus 0.4em\relax New York, NY, USA: Association for Computing Machinery, 2019, p. 151–158. [Online]. Available: \url{https://doi.org/10.1145/3308532.3329475}
\BIBentrySTDinterwordspacing

\bibitem{tapo2020is_more_autonomy}
T.~Bhattacharjee, E.~K. Gordon, R.~Scalise, M.~E. Cabrera, A.~Caspi, M.~Cakmak, and S.~S. Srinivasa, ``Is more autonomy always better? exploring preferences of users with mobility impairments in robot-assisted feeding,'' in \emph{2020 15th ACM/IEEE International Conference on Human-Robot Interaction (HRI)}, 2020, pp. 181--190.

\bibitem{integrativeorganizationaltrust}
\BIBentryALTinterwordspacing
R.~C. Mayer, J.~H. Davis, and F.~D. Schoorman, ``An integrative model of organizational trust,'' \emph{The Academy of Management Review}, vol.~20, no.~3, pp. 709--734, 1995. [Online]. Available: \url{http://www.jstor.org/stable/258792}
\BIBentrySTDinterwordspacing

\bibitem{weigelin2018trust}
B.~C. Weigelin, M.~Mathiesen, C.~Nielsen, K.~Fischer, and J.~Nielsen, ``Trust in medical human-robot interactions based on kinesthetic guidance,'' in \emph{2018 27th IEEE International Symposium on Robot and Human Interactive Communication (RO-MAN)}, 2018, pp. 901--908.

\bibitem{trustSAR}
L.~Beuscher, J.~Fan, N.~Sarkar, M.~Dietrich, P.~Newhouse, K.~Miller, and L.~Mion, ``Socially assistive robots: Measuring older adults' perceptions,'' \emph{Journal of Gerontological Nursing}, vol.~43, 07 2017.

\bibitem{robottherapist}
J.~Xu, D.~G. Bryant, and A.~Howard, ``Would you trust a robot therapist? validating the equivalency of trust in human-robot healthcare scenarios,'' in \emph{2018 27th IEEE International Symposium on Robot and Human Interactive Communication (RO-MAN)}, 2018, pp. 442--447.

\bibitem{robinette2016overtrust}
P.~Robinette, W.~Li, R.~Allen, A.~M. Howard, and A.~R. Wagner, ``Overtrust of robots in emergency evacuation scenarios,'' in \emph{2016 11th ACM/IEEE International Conference on Human-Robot Interaction (HRI)}, 2016, pp. 101--108.

\bibitem{robinette2017effect}
P.~Robinette, A.~M. Howard, and A.~R. Wagner, ``Effect of robot performance on human–robot trust in time-critical situations,'' \emph{IEEE Transactions on Human-Machine Systems}, vol.~47, no.~4, pp. 425--436, 2017.

\bibitem{morales2019interaction}
\BIBentryALTinterwordspacing
C.~G. Morales, E.~J. Carter, X.~Z. Tan, and A.~Steinfeld, ``Interaction needs and opportunities for failing robots,'' in \emph{Proceedings of the 2019 on Designing Interactive Systems Conference}, ser. DIS '19.\hskip 1em plus 0.5em minus 0.4em\relax New York, NY, USA: Association for Computing Machinery, 2019, p. 659–670. [Online]. Available: \url{https://doi.org/10.1145/3322276.3322345}
\BIBentrySTDinterwordspacing

\bibitem{king2010towards}
C.-H. King, T.~L. Chen, A.~Jain, and C.~C. Kemp, ``Towards an assistive robot that autonomously performs bed baths for patient hygiene,'' in \emph{2010 IEEE/RSJ International Conference on Intelligent Robots and Systems}, 2010, pp. 319--324.

\bibitem{kapusta2019system}
A.~S. Kapusta, P.~M. Grice, H.~M. Clever, Y.~Chitalia, D.~Park, and C.~C. Kemp, ``A system for bedside assistance that integrates a robotic bed and a mobile manipulator,'' \emph{Plos one}, vol.~14, no.~10, 2019.

\bibitem{Liu_2022_Bathing}
\BIBentryALTinterwordspacing
F.~Liu, V.~Patil, Z.~Erickson, and Z.~Temel, ``Characterization of a meso-scale wearable robot for bathing assistance,'' in \emph{2022 IEEE International Conference on Robotics and Biomimetics (ROBIO)}.\hskip 1em plus 0.5em minus 0.4em\relax IEEE, Dec. 2022. [Online]. Available: \url{http://dx.doi.org/10.1109/ROBIO55434.2022.10011741}
\BIBentrySTDinterwordspacing

\bibitem{madan2024rabbit}
R.~Madan, S.~Valdez, D.~Kim, S.~Fang, L.~Zhong, D.~T. Virtue, and T.~Bhattacharjee, ``Rabbit: A robot-assisted bed bathing system with multimodal perception and integrated compliance,'' in \emph{Proceedings of the 2024 ACM/IEEE International Conference on Human-Robot Interaction}, ser. HRI '24.\hskip 1em plus 0.5em minus 0.4em\relax New York, NY, USA: Association for Computing Machinery, 2024, p. 472–481.

\bibitem{feeding_tapo}
R.~K. Jenamani, D.~Stabile, Z.~Liu, A.~Anwar, K.~Dimitropoulou, and T.~Bhattacharjee, ``Feel the bite: Robot-assisted inside-mouth bite transfer using robust mouth perception and physical interaction-aware control,'' 2024.

\bibitem{feeding_dorsa}
P.~Sundaresan, J.~Wu, and D.~Sadigh, ``Learning sequential acquisition policies for robot-assisted feeding,'' 2023.

\bibitem{assistive_robots_review}
\BIBentryALTinterwordspacing
A.~Nanavati, V.~Ranganeni, and M.~Cakmak, ``Physically assistive robots: A systematic review of mobile and manipulator robots that physically assist people with disabilities,'' \emph{Annual Review of Control, Robotics, and Autonomous Systems}, vol.~7, no.~1, p. null, 2024. [Online]. Available: \url{https://doi.org/10.1146/annurev-control-062823-024352}
\BIBentrySTDinterwordspacing

\bibitem{kemp2022design}
C.~C. Kemp, A.~Edsinger, H.~M. Clever, and B.~Matulevich, ``The design of stretch: A compact, lightweight mobile manipulator for indoor human environments,'' 2022.

\bibitem{bazarevsky2020blazepose}
V.~Bazarevsky, I.~Grishchenko, K.~Raveendran, T.~Zhu, F.~Zhang, and M.~Grundmann, ``Blazepose: On-device real-time body pose tracking,'' 2020.

\bibitem{MeetObi_2023}
\BIBentryALTinterwordspacing
Dec 2023. [Online]. Available: \url{https://meetobi.com/meet-obi/\#whatisobi}
\BIBentrySTDinterwordspacing

\bibitem{NARS}
T.~Nomura, T.~Kanda, and T.~Suzuki, ``Experimental investigation into influence of negative attitudes toward robots on human-robot interaction,'' \emph{AI Soc.}, vol.~20, pp. 138--150, 03 2006.

\bibitem{NARS_Confirmation}
D.~S. Syrdal, K.~Dautenhahn, K.~Koay, and M.~Walters, ``The negative attitudes towards robots scale and reactions to robot behaviour in a live human-robot interaction study,'' 01 2009.

\bibitem{Madsen2000MeasuringHT}
\BIBentryALTinterwordspacing
M.~Madsen and S.~D. Gregor, ``Measuring human-computer trust,'' 2000. [Online]. Available: \url{https://api.semanticscholar.org/CorpusID:18821611}
\BIBentrySTDinterwordspacing

\bibitem{robot_trust_repair_review}
\BIBentryALTinterwordspacing
A.~L. Baker, E.~K. Phillips, D.~Ullman, and J.~R. Keebler, ``Toward an understanding of trust repair in human-robot interaction: Current research and future directions,'' \emph{ACM Trans. Interact. Intell. Syst.}, vol.~8, no.~4, nov 2018. [Online]. Available: \url{https://doi.org/10.1145/3181671}
\BIBentrySTDinterwordspacing

\bibitem{qualcoder}
\BIBentryALTinterwordspacing
Dec 2023. [Online]. Available: \url{https://github.com/ccbogel/QualCoder/releases/tag/3.5}
\BIBentrySTDinterwordspacing

\end{thebibliography}

\end{document}